\documentclass{article}

\usepackage[preprint]{style_file}

\usepackage[utf8]{inputenc} % allow utf-8 input
\usepackage[T1]{fontenc}    % use 8-bit T1 fonts
\usepackage{hyperref}       % hyperlinks
\usepackage{url}            % simple URL typesetting
\usepackage{booktabs}       % professional-quality tables
\usepackage{amsfonts}       % blackboard math symbols
\usepackage{nicefrac}       % compact symbols for 1/2, etc.
\usepackage{microtype}      % microtypography
\usepackage{xcolor}         % colors
\usepackage{amsmath}
\usepackage{graphicx}

\title{A Skull-Adaptive Framework for AI-Based 3D Transcranial Focused Ultrasound Simulation}

\author{
    \begin{tabular}{c}
          Vinkle Srivastav${^{1, 2}}$ \quad Juliette Puel$^{1, 2}$ \quad Jonathan Vappou$^1$ \quad Elijah Van Houten$^1$\\
         \quad Paolo Cabras$^{1, 3*}$ \quad Nicolas Padoy$^{1, 2*}$ \\
         \\
    \end{tabular}
    \\
  $^1$University of Strasbourg, CNRS, INSERM, ICube, UMR7357, Strasbourg, France \\
  $^2$IHU Strasbourg, Strasbourg, France \\
  $^3$Image Guided Therapy, Pessac, France\\  
  $^*$Shared last authorship 
}

\begin{document}
\maketitle

\begin{abstract}\label{sec:abstract}
Transcranial focused ultrasound (tFUS) is an emerging modality for non-invasive brain stimulation and therapeutic intervention, offering millimeter-scale spatial precision and the ability to target deep brain structures. However, the heterogeneous and anisotropic nature of the human skull introduces significant distortions to the propagating ultrasound wavefront, which require time-consuming patient-specific planning and corrections using numerical solvers for accurate targeting. To enable data-driven approaches in this domain, we introduce \emph{TFUScapes}, the first large-scale, high-resolution dataset of tFUS simulations through anatomically realistic human skulls derived from T1-weighted MRI images. We have developed a scalable simulation engine pipeline using the k-Wave pseudo-spectral solver, where each simulation returns a steady-state pressure field generated by a focused ultrasound transducer placed at realistic scalp locations.  In addition to the dataset, we present \emph{DeepTFUS}, a deep learning model that estimates normalized pressure fields directly from input 3D CT volumes and transducer position. The model extends a U-Net backbone with transducer-aware conditioning, incorporating Fourier-encoded position embeddings and MLP layers to create global transducer embeddings. These embeddings are fused with U-Net encoder features via feature-wise modulation, dynamic convolutions, and cross-attention mechanisms. The model is trained using a combination of spatially weighted and gradient-sensitive loss functions, enabling it to approximate high-fidelity wavefields. The \emph{TFUScapes} dataset is publicly released to accelerate research at the intersection of computational acoustics, neurotechnology, and deep learning. The dataset is available at \url{https://github.com/CAMMA-public/TFUScapes}.
\end{abstract}
\section{Introduction}\label{sec:Intro}
Transcranial focused ultrasound (tFUS) is an emerging technique for non-invasive therapeutic intervention~\cite{mahoney2023low,maloney2015emerging,hynynen2001noninvasive}. It enables energy delivery to deep brain targets with millimeter precision, opening up new frontiers in the treatment of neurological and psychiatric diseases~\cite{krishna2018review,kubanek2018neuromodulation,shi2025advances}. A fundamental challenge in tFUS lies in the propagation of acoustic waves through the human skull. Indeed, the skull - \emph{a highly heterogeneous, multilayered structure with varying density, sound speed, and sound attenuation} - acts as a complex acoustic lens that distorts ultrasound waves uniquely within each individual. To mitigate these issues and for assuring safe, efficient, and precise energy delivery to the intended target, it is essential to incorporate subject-specific acoustic modeling into the planning and delivery of tFUS interventions~\cite{treeby2010k,cueto2022stride}. 

Traditional computational methods for modeling ultrasound propagation use numerical solvers, such as finite element methods, finite difference methods, and spectral methods and solve the governing acoustic wave equations over discretized anatomical domains~\cite{cueto2022stride, samoudi2019computational, pasquinelli2020transducer}. These simulations incorporate subject-specific geometry and tissue properties derived from imaging modalities such as CT or MRI. While accurate, these numerical-solvers-based approaches are computationally intensive, often requiring several hours to simulate and optimize the procedure for a single case in 3D. This latency presents a substantial barrier for clinical workflows, especially in time-sensitive clinical settings.

To overcome these limitations, data-driven methods, particularly those based on deep learning, have been explored as a means of accelerating acoustic field estimation and correction~\cite{stanziola_helmholtz_2021,park2023real,naftchi2024deep}. These methods can learn to approximate complex mappings from anatomical data to ultrasound pressure fields, potentially enabling orders-of-magnitude speedups over traditional numerical solvers. However, the success of such approaches hinges on the availability of large-scale, high-quality datasets that accurately capture the ultrasound propagation across diverse anatomical structures. Currently available datasets and corresponding models are either restricted to 2D (with limited anatomical diversity)~\cite{stanziola_helmholtz_2021} or consist of small-scale 3D datasets that are not publicly accessible and specific to a single transducer geometry~\cite{park2023real}. As a result, there is a lack of open datasets that are both grounded in physical modeling and representative of realistic neuroanatomy. 

\begin{figure}[t]
    \centering
\includegraphics[width=0.95\columnwidth]{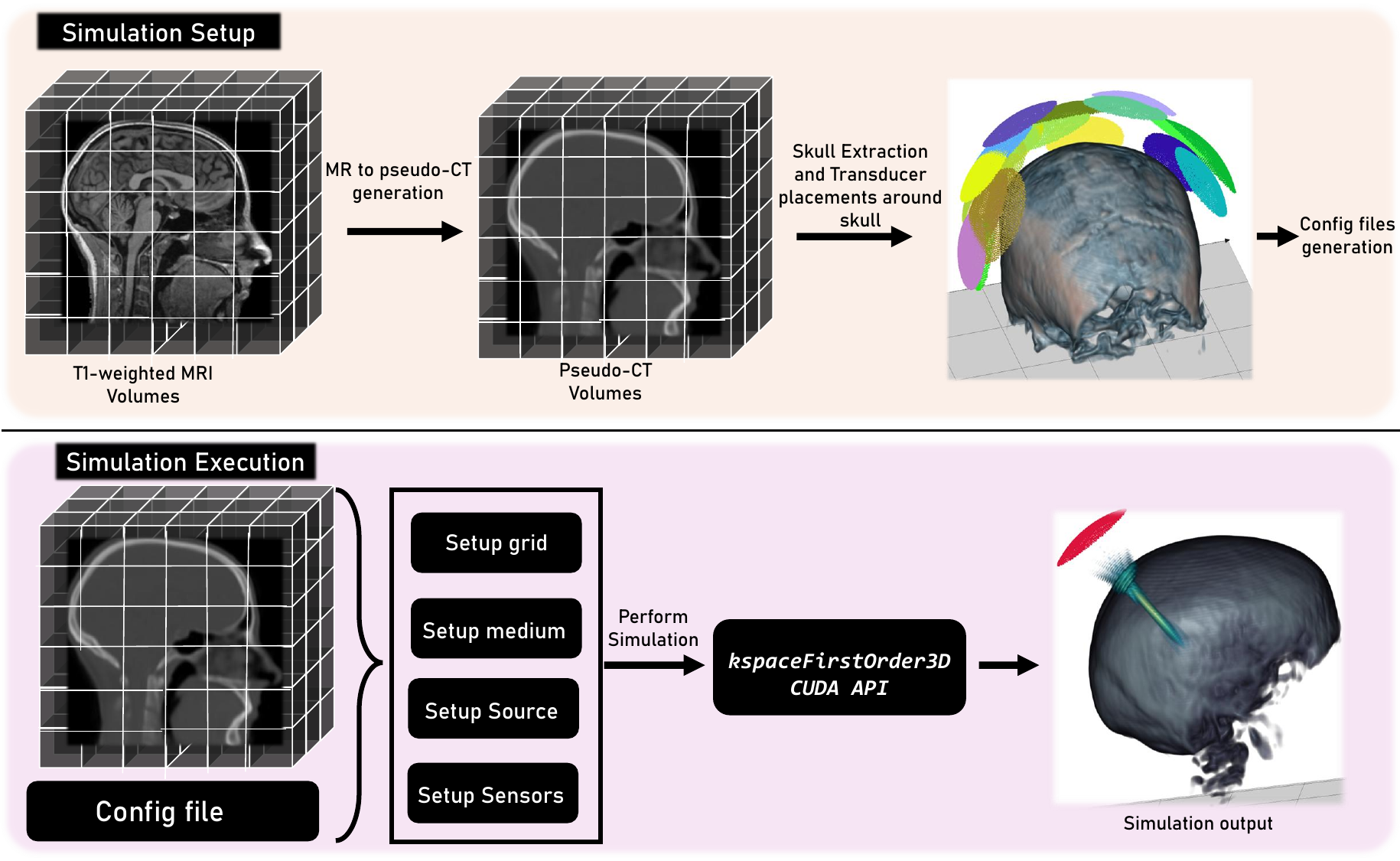}
    \caption{Overview of the \emph{TFUScapes} dataset generation pipeline for creating subject-specific transcranial focused ultrasound (tFUS) pressure fields using MRI-derived pseudo-CTs and GPU-accelerated k-Wave simulations.}
    \label{fig:TFUScapes}
\end{figure}

As our first contribution, we introduce \emph{TFUScapes}, the first large-scale dataset of anatomically realistic transcranial ultrasound simulations generated using a full-wave, pseudo-spectral acoustic solver. To construct this dataset, we have developed a scalable simulation pipeline built on top of the k-Wave library~\cite{treeby2010k,k-Wave-Python}, optimized for execution on GPU-based high-performance computing infrastructure. Each simulation is defined by a YAML configuration file that specifies subject-specific tissues' acoustic properties (namely density, sound speed, and absorption power law coefficients) to capture the complex acoustic heterogeneity of the skull~\cite{aubry2003experimental}. The configuration also includes transducer parameters such as position, orientation, focal length, and aperture, as well as emitted waves characteristics (fundamental frequency and amplitude). Each simulation outputs the 3D volumetric pressure field at steady state produced by a mono-element transducer positioned across multiple subjects at anatomically valid scalp locations. The final dataset includes $2,500$ simulations spanning $125$ subjects, with $20$ transducer placements per subject emitting a wave at a frequency of $500$ kHz. The dataset generation pipeline is illustrated in Figure~\ref{fig:TFUScapes}.

To complement our dataset and enable rapid inference of acoustic pressure fields, our second contribution is \emph{DeepTFUS} - a deep learning framework designed to predict normalized 3D acoustic pressure distributions from pseudo-CT volumes (generated from T1-weighted MRI scans~\cite{yaakub2023pseudo}) and transducer geometries. Unlike conventional medical image segmentation tasks, which involve mapping grid-structured inputs (e.g., CT or MRI volumes) to similarly structured outputs (e.g., segmentation maps)~\cite{ronneberger2015u}, our task requires predicting a grid-structured pressure field conditioned on both the input volume and a set of unordered 3D coordinates defining the transducer layout. To address this challenge, we extend the 3D U-Net architecture~\cite{ronneberger2015u} to extract spatial features from the pseudo-CT input. We introduce a \emph{transducer-aware conditioning module} that encodes the sparse transducer coordinates using high-frequency Fourier embeddings~\cite{tancik2020fourier}, producing a compact global representation of the transducer configuration. This representation is then fused into the U-Net pipeline through a combination of \emph{cross-attention modules}~\cite{vaswani2017attention}, \emph{feature-wise linear modulation (FiLM)}~\cite{perez2018film}, and \emph{dynamic convolutions}~\cite{chen2020dynamic}, enabling dynamic integration of anatomical and positional information throughout the network. The model architecture is shown in Figure~\ref{fig:DeepTFUS}.

Training of \emph{DeepTFUS} is guided by a composite loss function designed to address the inherent imbalance in the ground truth acoustic pressure maps, where most regions exhibit low pressure while high-pressure values are confined to small, localized areas. To ensure accurate prediction in these critical high-pressure zones, we employ a \emph{spatially weighted mean squared error (MSE)} that assigns greater importance to the regions of high pressure. This targeted weighting prevents the model from being dominated by the abundant low-pressure areas. Additionally, we incorporate a \emph{structural gradient-based loss} to promote the preservation of spatial detail and the smooth transitions characteristic of physically realistic pressure fields. Together, this loss formulation enables \emph{DeepTFUS} to reconstruct high-fidelity acoustic pressure distributions with improved accuracy in key regions, while also delivering substantial improvements in computational efficiency and scalability.

By publicly releasing the \emph{TFUScapes} dataset, we aim to accelerate research at the intersection of computational acoustics, neurotechnology, and deep learning. %This work provides a foundation for future data-driven approaches in tFUS planning and real-time field estimation, paving the way toward clinically deployable and efficient non-invasive brain stimulation systems.

\section{Related work}

%Overview of the scalable pipeline used to generate the TFUScapes dataset for tFUS modeling. T1-weighted MRI volumes are converted into pseudo-CT volumes and paired with randomized transducer placements. Each setup is run using GPU-accelerated k-Wave simulations with anatomically derived acoustic parameters. The resulting dataset contains 3D pressure fields aligned with subject-specific anatomy and transducer geometry configurations.
\subsection{transcranial Focused Ultrasound (tFUS)}
Ultrasound is a widely used imaging modality due to its portability, safety, and cost-effectiveness. Beyond its diagnostic applications, Focused Ultrasound (FUS) has emerged as a powerful therapeutic tool with diverse clinical applications, including tumor ablation via coagulative necrosis~\cite{maloney2015emerging, cabras2022new}, blood-brain barrier disruption~\cite{hynynen2001noninvasive}, neuromodulation~\cite{mahoney2023low}, and tissue regeneration~\cite{ricotti2024ultrasound}. FUS's therapeutic efficacy and safety depend critically on the precise delivery of acoustic energy to the intended target. This requirement in transcranial FUS (tFUS) is particularly stringent due to the skull's complex structure and heterogeneous acoustic properties, where variations in skull geometry, impedance and attenuation can cause beam distortion, resulting in focal point shifts or off-target energy deposition, which may compromise treatment effectiveness and patient safety~\cite{sigona2024ultrasound}. tFUS treatment relies not only on accurate transducer placement but also on robust modeling of acoustic wave propagation through the skull and brain. Currently, preoperative planning for tFUS involves iterative numerical simulations based on patient-specific anatomical models~\cite{aubry2003experimental}. These simulations aim to predict wave behavior and optimize sonication parameters but are often slow, taking several hours, and are heavily reliant on manual tuning and operator expertise. This workflow limits scalability and presents a major bottleneck to broader clinical adoption.  

\subsection{Computational modeling for tFUS}
A variety of numerical methods have been developed to simulate acoustic wave propagation in biological tissues, particularly in the context of transcranial focused ultrasound (tFUS). Classical approaches such as finite element methods, finite difference methods, and spectral methods have been widely employed~\cite{cueto2022stride, samoudi2019computational, pasquinelli2020transducer}. Among these, pseudo-spectral time-domain (PSTD) methods, particularly those implemented in the k-Wave toolbox~\cite{treeby2010k,treeby2012modeling}, have become a \textit{de facto} standard for modeling ultrasound propagation through highly heterogeneous media such as the human skull, due to their high accuracy and ability to model nonlinear effects and frequency-dependent absorption with fewer spatial and temporal resolution. To improve simulation speed, hybrid strategies, e.g. combining Rayleigh-integral with the k-Wave library, reduce domain size by modeling transducer fields analytically in water, but still face high costs due to k-Wave's fine resolution needs~\cite{rosnitskiy_simulation_2019}. Faster alternatives like hybrid angular spectrum methods and ray tracing offer efficiency in layered or smoothly varying media, yet struggle with pressure estimation accuracy in highly scattering or anatomically complex regions~\cite{vyas_ultrasound_2012,bancel_comparison_2021}. Deep learning has started to offer a promising alternative by learning end-to-end mappings from anatomical features to acoustic fields~\cite{stanziola_helmholtz_2021,park2023real,naftchi2024deep}. Once trained, these vmodels can deliver near-instantaneous predictions while maintaining a level of accuracy comparable to conventional physics-based simulations. This speed-up could not only make large-scale simulation feasible but also open the door to real-time treatment planning, optimization, and uncertainty quantification, accelerating the clinical adoption of tFUS.

\section{Methodology}\label{sec:methods}

\subsection{Dataset construction: \emph{TFUScapes}}\label{sec:dataset}

To enable high-fidelity, data-driven modeling of transcranial focused ultrasound (tFUS), we have developed a scalable simulation pipeline using the k-Wave library~\cite{treeby2010k,k-Wave-Python} and YAML-based configuration files~\cite{ben2009yaml}, designed for parallel execution on GPU-powered high-performance computing systems. The simulation generation contains two main steps: \emph{1) simulation setup} - converting T1-weighted MRI scans to pseudo-CTs and constructing config files for the simulation, \emph{2) simulation execution} - running GPU-accelerated k-Wave simulations with subject-specific acoustic properties to build the \emph{TFUScapes} dataset. 

\paragraph{Simulation setup.} We acquire $125$ manually skull-stripped T1-weighted anatomical MRI scans from the NFBS dataset~\cite{puccio2016preprocessed}. These scans are converted into pseudo-CT volumes using the MR-to-pCT model to generate anatomically realistic head models suitable for acoustic simulation~\cite{yaakub2023pseudo}. Then, for each subject, a bowl-shaped transducer is placed at $20$ clinically-relevant and randomly selected positions on the scalp. Also, the transducer geometry is randomly defined within certain limits: both the focal length and the aperture diameter are between $55$ mm and $75$ mm. This results in $2,500$ unique simulation configurations ($125$ $\times$ $20$ = $2,500$), each defined by the subject-specific anatomy, source parameters, and the spatial configuration of the acoustic source with respect to the scalp. %All simulations are designed to be parallelizable and to be executed on a high-performance GPU cluster to enable scalable and efficient data generation. 

\paragraph{Simulation execution.} For each simulation configuration, we define a 3D grid $\mathbf{x}\in\mathbb{R}^{N_x\times N_y\times N_z}$ with spacing, $\Delta x = \Delta y = \Delta z = \frac{c_0}{f_0 \cdot \mathrm{ppw}}$, where $c_0$ is the reference sound speed, $\mathrm{ppw}$ is the number of points per wavelength, and $f_0$ is the ultrasound transducer frequency, and $N_x, N_y, N_z$ is the 3D grid size. The temporal resolution, $\Delta t$, is determined by a Courant–Friedrichs–Lewy (CFL) condition~\cite{de2013courant} to ensure numerical stability: $\Delta t \le \frac{\Delta x}{\sqrt{3}\,c_{0}}$.

{\hskip 2em} To define the acoustic medium, we resample the pseudo-CT volume to $0.5$ mm isotropic voxel-spacing and convert to spatially varying acoustic parameters: density $\rho(x)$, sound speed $c(x)$, and attenuation $\alpha(x)$. Here $x\in\mathbb{R}^{3}$ is a 3D coordinate in the simulation domain. These parameters are derived from Hounsfield units (HU) using the following empirical mappings~\cite{yaakub2023pseudo, BRIC_TUS_Simulation_Tools}:
\begin{subequations}
\label{eq:ct_mapping}
\begin{align}
\rho(x) &= \rho_{\min} + (\rho_{\max}-\rho_{\min})\,\frac{\mathrm{HU}(x)}{\mathrm{HU}_{\max}},\\
c(x) &= c_{\min} + (c_{\max}-c_{\min})\,\frac{\rho(x)-\rho_{\min}}{\rho_{\max}-\rho_{\min}},\\
\alpha(x) &= \alpha_{\min} + (\alpha_{\max}-\alpha_{\min})\,\Biggl(1 - \sqrt{\frac{\mathrm{HU}(x)-\mathrm{HU}_{\min}}{\mathrm{HU}_{\max}-\mathrm{HU}_{\min}}}\Biggr).
\end{align}
\end{subequations}

{\hskip 2em} The source is modeled as a 3D concave bowl transducer with specified diameter and radius of curvature. The acoustic pressure field is driven by a continuous wave (CW) sinusoidal source: 
\begin{equation}
p(t) = A \cos(2\pi f_0 t + \phi),
\label{eq:transducer}
\end{equation}
where $A$ is the amplitude and $\phi$ is the initial phase. The pressure source is defined via the \texttt{kWaveArray} API and projected onto the simulation grid as a 3D mask. Each simulation runs to steady state using the CUDA-accelerated \texttt{kspaceFirstOrder3D} solver, which models the coupled first-order acoustic wave equations in heterogeneous media. After simulation, time-series pressure data are post-processed slab-by-slab using the Fourier transform, $\mathcal{F}$, to extract the amplitude at the driving frequency $f_0$:
\[
\text{Amplitude}  = P(f_0) = \mathcal{F}\{p(t)\}\bigl|_{f_0}.
\]

The final dataset, named \emph{TFUScapes}, comprises volumetric pressure maps, along with pseudo-CT volumes and metadata detailing transducer configuration and subject-specific acoustic properties.  After the simulation, we extract a $256 \times 256 \times 256$ subvolume centered on the region of interest, which includes the transducer and the portion of the skull exhibiting the maximum pressure envelope. The cropped pseudo-CT, transducer geometry, and the pressure maps are saved in a compressed numpy \texttt{.npz} file. The dataset generation pipeline is illustrated in Figure~\ref{fig:TFUScapes}. The released dataset comprises simulation examples, each stored in a separate \texttt{.npz} file. Each file contains a cropped pseudo-CT volume, the corresponding pressure map, and the 3D coordinates of the transducer geometry.

\subsection{Architecture: \emph{DeepTFUS}}\label{sec:surr_model}
\begin{figure}[t]
    \centering
\includegraphics[width=1.0\columnwidth]{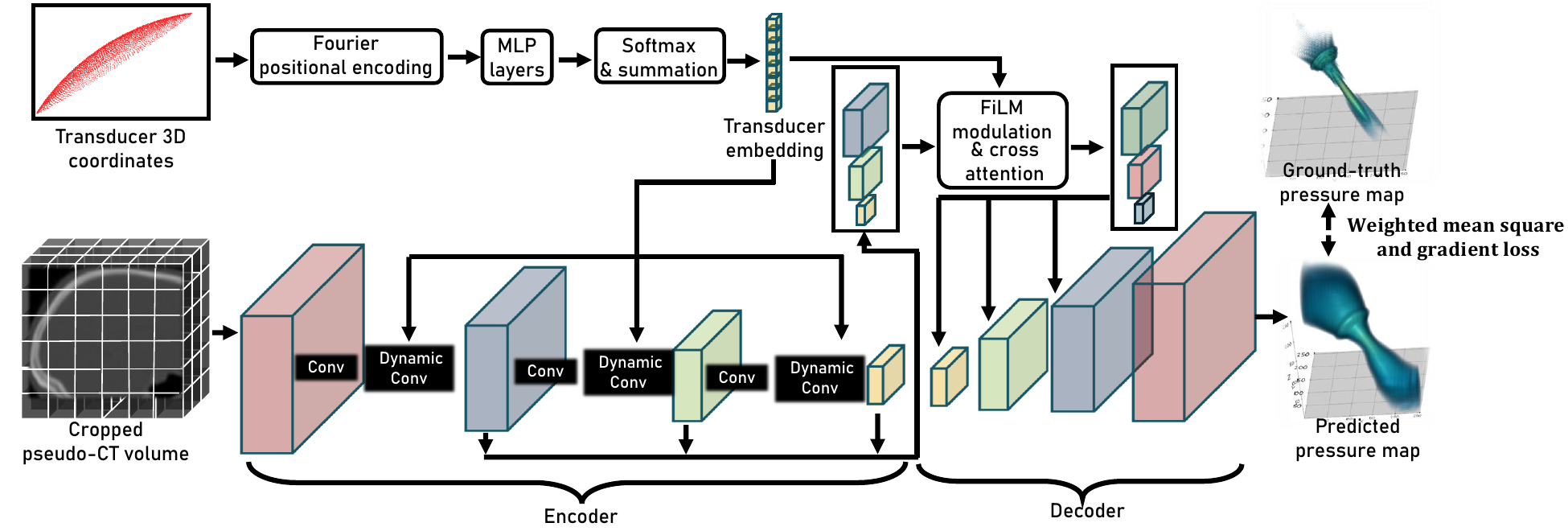}
    \caption{Overview of the proposed transducer-aware 3D U-Net framework, \emph{DeepTFUS}, for predicting tFUS pressure fields from pseudo-CT anatomy and transducer geometry.}
    \label{fig:DeepTFUS}
\end{figure}
%The model uses a 3D U-Net architecture with transducer-aware feature fusion. Transducer geometry is encoded via Fourier positional encodings and attention-based pooling into a global vector. This embedding conditions the U-Net through dynamic convolutions and FiLM modulation in the encoder to capture both anatomical features from pseudo-CT and transducer representation from sparse 3D coordinates.
Given the \emph{TFUScapes} dataset, we propose a deep learning framework for predicting 3D tFUS pressure fields from patient-specific pseudo-CTs and transducer configurations. Each training sample consists of: (1) a high-resolution cropped pseudo-CT volume  $\mathbf{X} \in \mathbb{R}^{256 \times 256 \times 256}$; (2) a set of coordinates of the active surface from the transducer (emitting the acoustic wave) $\mathbf{T} = \{\mathbf{t}_i\}_{i=1}^{N} \subset \mathbb{R}^3$, where $N$ is the number of points sampled from the bowl-shaped active surface and each $\mathbf{t}_i$ denotes its 3D coordinate in physical space; and (3) a target pressure field $\mathbf{P} \in \mathbb{R}^{256 \times 256 \times 256}$, representing the normalized peak pressure amplitude $p_{\text{max}}$ at each voxel in the simulation domain.

The model architecture is based on a 3D U-Net with an encoder-decoder structure and skip connections~\cite{ronneberger2015u}. To enable the model to incorporate transducer configuration as a conditioning signal, we introduce a feature extraction pipeline from the transducer active surface coordinates, which embeds their spatial structure into a global vector representation. The coordinates \( \mathbf{t}_i \) are first normalized to the \([0, 1]^3\) range and passed through a high-frequency Fourier positional encoding:
\begin{equation}
\gamma(\mathbf{t}_i) = \left[\sin(2^0 \pi \mathbf{t}_i), \cos(2^0 \pi \mathbf{t}_i), \ldots, \sin(2^{n-1} \pi \mathbf{t}_i), \cos(2^{n-1} \pi \mathbf{t}_i)\right],
\end{equation}

which captures multiscale geometric detail. These embeddings are concatenated with the original coordinates and processed via MLP layers, followed by layer normalization and ReLU activations to produce point-wise features, $\mathbf{z}_i$. To aggregate the individual point features into a fixed-size representation for each volume, we employ attention-based weighted pooling. For each embedding $\mathbf{z}_i$, we compute a scalar attention weight: $\alpha_i = \frac{\exp(w^\top \mathbf{z}_i)}{\sum_j \exp(w^\top \mathbf{z}_j)}$, where $w$ is a learned projection vector. The final embedding \( \mathbf{z}_T \in \mathbb{R}^d \) is computed as the weighted sum over all embeddings: $\mathbf{z}_T = \sum_{i} \alpha_i \mathbf{z}_i$, providing a permutation-invariant representation of the transducer geometry. The encoded transducer embedding, $\mathbf{z}_T$, is integrated into the U-Net pipeline in three distinct ways.

{\hskip 2em} First, during the encoding path of the U-Net, we employ \textit{Dynamic Convolutions}~\cite{chen2020dynamic} that modulate the encoder layers using kernels generated conditionally from \( \mathbf{z}_T \). These dynamic kernels are predicted by an MLP and applied via grouped depthwise 3D convolution to modulate the U-Net intermediate feature maps.

{\hskip 2em} Second, the transducer embedding is used in the decoding path via \textit{Feature-wise Linear Modulation (FiLM)}~\cite{perez2018film}. The transducer embedding $\mathbf{z}_T$ is linearly projected to obtain a pair of scale and shift parameters, $ \gamma, \beta$, which are reshaped and applied to modulate the U-Net intermediate feature maps, $ \mathbf{f}$, as $\text{FiLM}(\mathbf{f}) = \gamma \cdot \mathbf{f} + \beta$.
This enables the decoder to condition its reconstruction of the pressure field on transducer-specific parameters. 

{\hskip 2em} Third, we implement \textit{cross-attention modules}~\cite{vaswani2017attention} at each encoder level to explicitly exchange information between the CT features and the transducer embedding. At each spatial resolution, the transducer embedding \( \mathbf{z}_T \) is used as a global token that attends to the spatially flattened CT features (treated as tokens), and vice versa. These bi-directional attention modules consist of two multi-head attention blocks: one from the transducer embedding to CT features and the other from CT features to the transducer token. The updated CT features are reshaped back to 3D and fused with the broadcasted transducer features through concatenation and \(1 \times 1 \times 1\) convolutional projection. 

These three design choices ensure that the encoder features integrate global transducer context with local anatomical information. These transducer-aware encoder features are then integrated with the U-Net decoder to produce the output pressure map $\mathbf{\hat{P}}$. The model is trained using a composite loss function that captures both voxel-wise accuracy and structural fidelity of the predicted pressure fields. The first component is a \textit{weighted mean squared error loss}:
\begin{equation}
    \mathcal{L}_{\text{weighted}} = \frac{1}{|\Omega|} \sum_{\mathbf{v} \in \Omega} w(\mathbf{v}) \|\mathbf{\hat{P}}(\mathbf{v}) - \mathbf{P}(\mathbf{v})\|^2,
    \label{eq:weighted_l2}
\end{equation}
where, $\mathbf{v}$ is the 3D coordinate and \( \Omega \) indexes all voxels in the volume, and the weight, $w(\mathbf{v})$, is defined as
\begin{equation}
w(\mathbf{v}) = \frac{\exp\left( \alpha \left( \mathbf{P}(\mathbf{v}) - \max_{\mathbf{v}'} \mathbf{P}(\mathbf{v}') \right) \right)}{\mathbb{E}_{\mathbf{v}} \left[ \exp\left( \alpha \left( \mathbf{P}(\mathbf{v}) - \max_{\mathbf{v}'} \mathbf{P}(\mathbf{v}') \right) \right) \right]}.
\label{eq_w}
\end{equation}
This exponentially weighted loss prioritizes high-pressure regions, which are clinically and physically more significant but spatially localized to a small region. The second component is a \textit{gradient consistency loss} that enforces alignment between the spatial derivatives of the prediction and ground truth:
\begin{equation}
\mathcal{L}_{\text{grad}} = \frac{1}{3} \sum_{i \in \{x, y, z\}} \| \nabla_i \hat{\mathbf{P}} - \nabla_i \mathbf{P} \|_2^2,
\label{eq:grad_l2}
\end{equation}
where \( \nabla_i \) denotes the finite-difference approximation of the spatial derivative along axis \( i \in \{x, y, z\} \). The final loss is defined as $\mathcal{L} = \mathcal{L}_{\text{weighted}} + \lambda \mathcal{L}_{\text{grad}}$, where $\lambda$ is a tunable hyperparameter. This loss encourages the model to preserve spatial coherence, especially at the boundaries of focal zones and in regions with high pressure gradients. The overall architecture is shown in Figure~\ref{fig:DeepTFUS}.

\section{Experiments}\label{sec:exp}

\subsection{Implementation details.}\label{sec:implementation}
\textbf{Simulation configuration and dataset generation.}\label{sec:exp_dataset}
The configuration files for the simulation were generated using a Python script, which populates a Jinja2 template to create subject-specific YAML files. For each subject, $20$ configurations are generated by sampling source positions and bowl transducer geometries. The bowl's radius of curvature (ROC) and diameter are randomly sampled within $\pm 10\,\text{mm}$ of a base value of $65\,\text{mm}$, yielding a range of $[55, 75]\,\text{mm}$ for both. The points per wavelength $\mathrm{ppw}$ is set to  $6$ and the reference sound speed $c_0$ is set to $1500$ m/s. This yields a spatial resolution of $\Delta x = \Delta y = \Delta z = 0.5 \text{mm}$ (being the fundamental frequency at $500$ kHz). The physical dimensions of the simulation grid were derived from the canonical pseudo-CT image (with $0.5$ mm voxel spacing) scaled by $\Delta x$. Source position offsets in the $x$, $y$, and $z$ directions were randomly selected from a uniform range of $[40, 60]$ voxels, with an additional fixed offset of $10$ voxels on each side to ensure boundary padding. The actual transducer position was randomized within these bounds, making sure it is pointing towards the center of the simulation grid. The simulation grid included additional padding using a perfectly matched layer (PML), whose optimal size was computed automatically for stability.

{\hskip 2em}Each generated configuration was used in the simulation pipeline which invokes a GPU-based k-Wave-based 3D acoustic solver - \texttt{kspaceFirstOrder3D}. The time step $\Delta t$ was determined by the CFL condition, using $\Delta t = 1 / (\texttt{ppp} \times f_0)$, where \texttt{ppp} (points-per-period) is derived from \texttt{ppw} and a fixed CFL number (typically $0.3$). The number of time steps $N_t$ was computed from the total simulation time $t_\text{end}$, and the final pressure field was recorded over the last few acoustic cycles. The acoustic medium was constructed from the CT data, with material properties (density $\rho(x)$, speed of sound $c(x)$, absorption $\alpha(x)$) computed based on linear or square-root mappings from Hounsfield units. The following values are used in the Equation~\ref{eq:ct_mapping}: $\rho_{\min} = 1000$ kg/m³, $\rho_{\max} = 1900$ kg/m³, $c_{\min} = 1500$ m/s, $c_{\max} = 3100$ m/s, $\alpha_{\min} = 4$ dB/(MHz cm), $\alpha_{\max} = 8.7$ dB/(MHz cm), with $\mathrm{HU}_{min} = 300$ and $\mathrm{HU}_{max} = 2000$. Soft tissues and brain matter (typically around $ HU\approx 0$) are thus modeled approximately as water. The acoustic source was modeled as a bowl-shaped transducer with a continuous-wave excitation of $f_0 = 500\,\text{kHz}$, amplitude of $A = 60000$, and a phase offset of $\phi= 0^\circ$, modulated by a cosine envelope (in the Equation~\ref{eq:transducer}). Running a full simulation (including data processing and saving results as numpy arrays) takes around 30 minutes per instance on an NVIDIA A100 GPU. 

\textbf{Model training.}\label{sec:exp_model_train}
The model is trained on a single NVIDIA A100 GPU with a DRAM of $80$GB for $50$ epochs using the AdamW optimizer~\cite{loshchilov2017decoupled}, which applies decoupled weight decay regularization. Training is performed with a batch size of $4$, and the initial learning rate is set to $1 \times 10^{-3}$. A cosine annealing schedule is used to gradually decrease the learning rate during training. The weighting factor $\alpha$ in Equation~\ref{eq_w} is set to $5.0$, and the weighting factor $\lambda$ in the total loss function $\mathcal{L}$ is set to $0.1$. During training, the CT volume is normalized using dynamic min-max normalization, and the pressure map is first divided by the max pressure and then log-transformed as $log(1 + x)$ to compress the dynamic range. Out of $125$ subjects, we randomly choose $85$ subjects for training, $10$ subjects for validation, and $30$ subjects for testing. 

\textbf{Metrics.}\label{sec:exp_metrics}
Let $\mathbf{P} \in \mathbb{R}^N$ denote the normalized ground truth pressure field and $\hat{\mathbf{P}} \in \mathbb{R}^N$ the predicted field, where $N = 256 \times 256 \times 256$ is the total number of voxels. We evaluate predictions using three metrics. First, we compute the \textbf{relative\_l2} as $\sqrt{ \| \hat{\mathbf{P}} - \mathbf{P} \|_2^2 / \| \mathbf{P} \|_2^2 }$. Since our goal is to localize the region of maximum pressure, we also adopt two metrics from the benchmark study by Aubry et al.~\cite{aubry2022benchmark}. Let $\mathbf{r}_\text{gt} \in \mathbb{R}^3$ and $\mathbf{r}_\text{pred} \in \mathbb{R}^3$ be the coordinates of the maximum values in $\mathbf{P}$ and $\hat{\mathbf{P}}$, respectively. The \textbf{focal\_position\_error} is defined as $\| \mathbf{r}_\text{gt} - \mathbf{r}_\text{pred} \|_2$. Let $P_{\text{max}} = \max(\mathbf{P})$ and $\hat{P}_{\text{max}} = \max(\hat{\mathbf{P}})$ be the maximum normalized pressure value in the ground truth and the prediction. Then, \textbf{max\_pressure\_error} is given by $100 \times \left| \hat{P}_{\text{max}} - P_{\text{max}} \right| / P_{\text{max}}$. The median, mean, and standard deviation are calculated across all the test samples. 

% \begin{table}[t!]
%     \centering
%     \resizebox{0.8\linewidth}{!}{%
%     \renewcommand{\arraystretch}{1.6}
%     \begin{tabular}{lll}
%         \toprule
%         \textbf{Metric} & \textbf{Definition} & \textbf{Explanation} \\
%         \midrule
%         Max pressure error & 
%         $ \frac{|\max (p_1) - \max (p_2)|}{\max (p_1)} $ & 
%         Relative error in maximum pressure amplitude \\
        
%         Focal position error & 
%         $ \| \arg\max(p_1) - \arg\max(p_2) \|_2 $ & 
%         Euclidean distance between focal points \\
        
%         Focal volume (-6dB) & 
%         $|V_1(-6\text{dB}) - V_2(-6\text{dB})|$ & 
%         Difference in focal volume using 50\% threshold \\
        
%         Focal volume (-3dB) & 
%         $|V_1(-3\text{dB}) - V_2(-3\text{dB})|$ & 
%         Difference in focal volume using 70\% threshold \\
        
%         Lateral FWHM error & 
%         $|x_1 - x_2|$ & 
%         Difference in lateral width of -6dB focal region \\
        
%         Axial FWHM error & 
%         $|z_1 - z_2|$ & 
%         Difference in axial length of -6dB focal region \\
        
%         IoU & 
%         $ \frac{V_1(-6\text{dB}) \cap V_2(-6\text{dB})}{V_1(-6\text{dB}) \cup V_2(-6\text{dB})} $ & 
%         Overlap ratio of focal volumes at -6dB \\
%         \bottomrule
%     \end{tabular}
%     } % end resizebox
%     \caption{Summary of comparison metrics used to evaluate 3D acoustic pressure fields. $p_1$ and $p_2$ are pressure amplitudes from two simulations.}
%     \label{tab:metrics}
% \end{table}

\subsection{Experiments}

\begin{table}[t!]
\centering
\caption{Ablation study of loss functions and architectural components on transcranial pressure field prediction performance. (median and mean ± std, bold = best)}
\label{tab:results}
\resizebox{\textwidth}{!}{
\begin{tabular}{lllllll}
\toprule
metric & \multicolumn{2}{c}{\textbf{relative\_l2}} & \multicolumn{2}{c}{\textbf{focal\_position\_error}} & \multicolumn{2}{c}{\textbf{max\_pressure\_error}} \\
 & mean ± std & median & mean ± std & median & mean ± std & median \\
  &  & & [mm] & [mm] & [\%] & [\%] \\
experiments &  &  &  &  &  &  \\
\midrule
Loss $\mathcal{L}_{\text{2}}$ & $\mathbf{40.25 \pm 6.80}$ & $\mathbf{39.21}$ & $5.49 \pm 3.72$ & $4.64$ & $92.39 \pm 30.91$ & $93.20$ \\
Loss $\mathcal{L}_{\text{weighted}}$ & $41.71 \pm 7.89$ & $39.98$ & $3.04 \pm 2.07$ & $2.55$ & $22.74 \pm 16.46$ & $19.88$ \\
\midrule
No dynamic conv. & $44.08 \pm 7.28$ & $42.47$ & $2.98 \pm 2.03$ & $2.60$ & $22.71 \pm 16.75$ & $20.65$ \\
No Fourier enc. & $42.82 \pm 8.98$ & $40.57$ & $3.14 \pm 2.26$ & $2.69$ & $18.47 \pm 14.64$ & $15.31$ \\
No FiLM & $41.31 \pm 8.62$ & $39.71$ & $3.11 \pm 2.25$ & $2.55$ & $\mathbf{14.65 \pm 12.59}$ & $\mathbf{12.25}$ \\
\midrule
\emph{DeepTFUS}$_{\text{tiny}}$ & $41.02 \pm 7.58$ & $39.40$ & $2.95 \pm 2.23$ & $2.55$ & $19.64 \pm 15.39$ & $17.03$ \\
\midrule
\emph{DeepTFUS} & $41.40 \pm 8.62$ & $39.37$ & $\mathbf{2.89 \pm 2.14}$ & $\mathbf{2.45}$ & $19.90 \pm 15.79$ & $16.63$ \\
\bottomrule
\end{tabular}
}
\end{table}
\begin{figure}[t!]
    \centering
    \includegraphics[width=1.0\columnwidth]{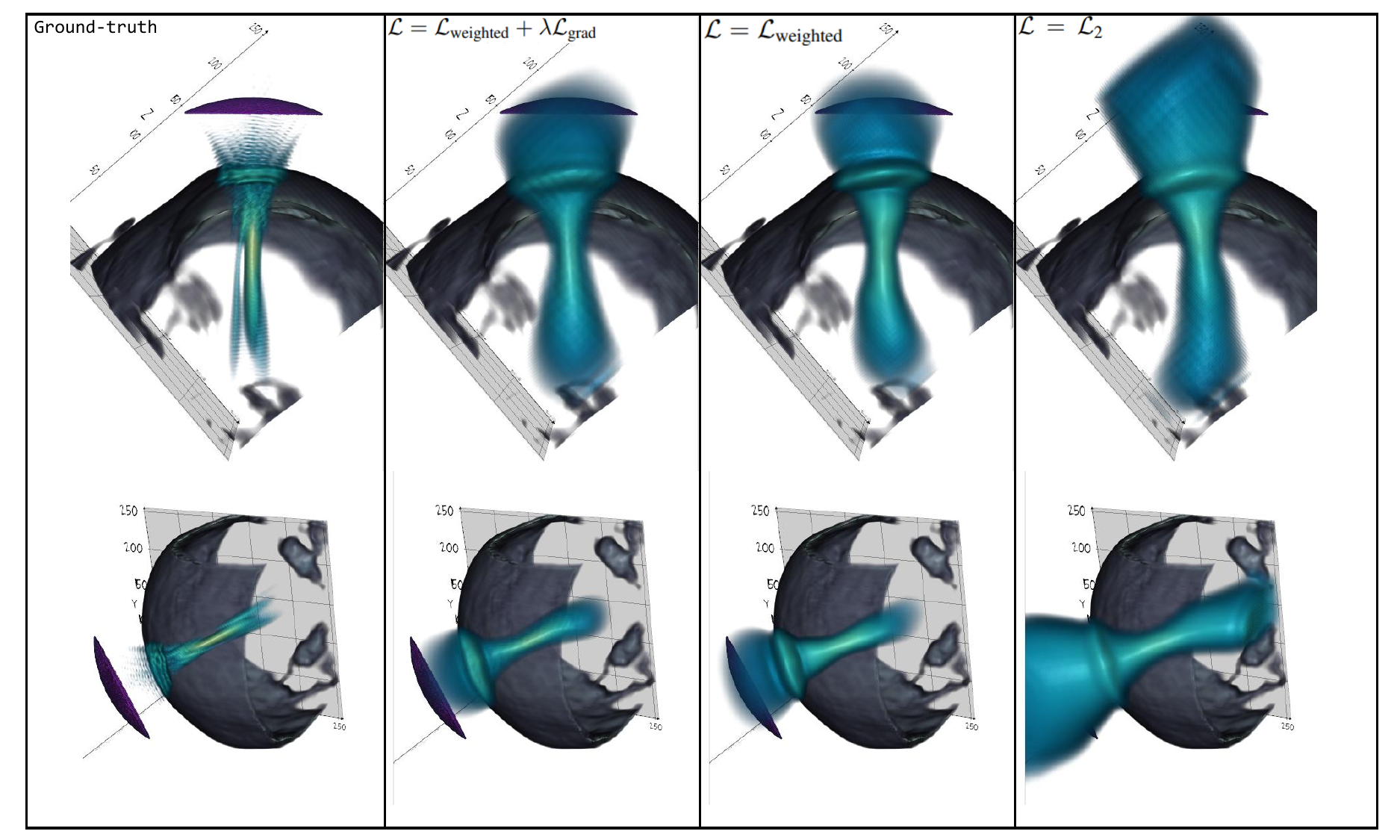}
    \caption{Qualitative results comparing different loss functions.}
    \label{fig:ablation_Loss}
\end{figure}
\begin{figure}[t!]
    \centering
    \includegraphics[width=1.0\columnwidth]{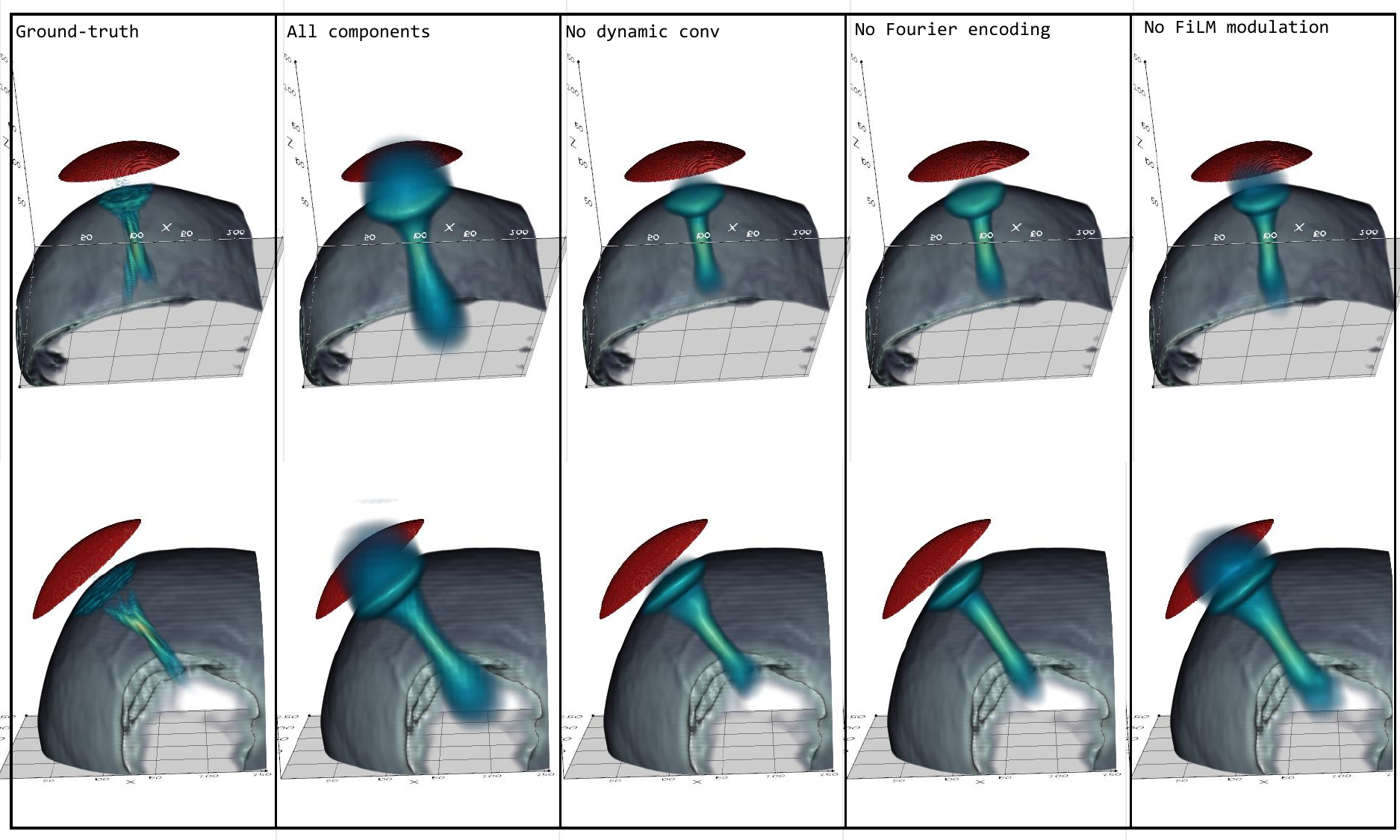}
    \caption{Qualitative results of different architecture choices.}
    \label{fig:ablation_arch}
\end{figure}

To evaluate the effectiveness of different model design choices, we conducted a series of experiments covering loss functions and architectural components. First, we compared two loss functions: the standard $\mathcal{L}_2$ loss and a weighted loss $\mathcal{L}_{\text{weighted}}$ that emphasizes high-pressure regions to better guide learning toward focal areas. Next, we investigated the contribution of specific architectural modules by disabling them one at a time: dynamic convolutions (responsible for spatially adaptive filtering conditioned on the transducer embedding), Fourier postional encoding (which enhances spatial localization via high-frequency positional information of transducer coordinates), and the FiLM (Feature-wise Linear Modulation) module, which provides transducer-aware feature conditioning. Finally, we evaluated two full-model variants: \emph{DeepTFUS}$_\text{tiny}$, which uses a smaller U-Net backbone for efficiency, and the full \emph{DeepTFUS} model, which incorporates all components and uses the weighted loss in combination with a gradient consistency loss given in Equations~\ref{eq:weighted_l2} and ~\ref{eq:grad_l2}.
\subsection{Results}  

Table~\ref{tab:results} presents the results across all metrics: relative\_l2, focal\_position\_error, and max\_pressure\_error. The weighted loss significantly improves localization performance compared to the standard $\mathcal{L}_2$ loss, reducing focal position error from $5.49 \pm 3.72$ to $3.04 \pm 2.07$, and maximum pressure error from $92.39 \pm 30.91$ to $22.74 \pm 16.46$, while maintaining similar global accuracy. Among the architectural ablations, removing dynamic convolutions slightly degrades global accuracy but improves localization. Removing Fourier encoding has a minor impact on performance, indicating modest reliance on positional encodings. Interestingly, disabling the FiLM module leads to the lowest maximum pressure error ($14.65 \pm 12.59$), suggesting it may introduce some overfitting in amplitude prediction despite helping spatial conditioning. The \emph{DeepTFUS}$_\text{tiny}$ model performs competitively, striking a good balance between accuracy and efficiency. The full \emph{DeepTFUS} model achieves the best focal position accuracy ($2.89 \pm 2.14$) and strong maximum pressure estimation. Figure~\ref{fig:ablation_Loss} and \ref{fig:ablation_arch} show some qualitative results comparing different loss functions and architecture choices. More qualitative results, including the failure cases, are shown in the supplementary material.

% \begin{table}[t!]
% \centering
% \caption{Ablation Studies}
% \label{Tab:Ablation}
% \begin{tabular}{lcccccccl}\toprule
% & \multicolumn{4}{c}{} & \multicolumn{4}{c}{$Surg-RET$}
% \\\cmidrule(lr){6-9}
%          COP & CTX & FTM & MLM & VE & R@1 & R@5 & R@10 & Average \\\midrule
% $\times$ & $\times$ & $\times$ & $\checkmark$ & $\checkmark$  & $9.2_{\pm0.7}$ & $32.4_{\pm2.9}$ & $47.2_{\pm3.9}$ & $29.6_{\pm2.5}$ \\\midrule
% $\checkmark$ & $\times$ & $\times$ & $\checkmark$ & $\checkmark$ & $11.7_{\pm1.4}$ & $37.1_{\pm2.3}$ & $50.5_{\pm5.7}$ & $33.1_{\pm3.1}$\\
% $\checkmark$ & $\checkmark$ & $\times$ & $\checkmark$ & $\checkmark$ & $12.9_{\pm1.0}$ & $38.9_{\pm0.4}$ & $56.1_{\pm1.6}$ & $36.0_{\pm1.0}$\\
% $\checkmark$ & $\checkmark$ & $\checkmark$ & $\checkmark$ & $\checkmark$  & $13.5_{\pm0.9}$ & $41.9_{\pm3.0}$ & $57.3_{\pm4.1}$ & $37.6_{\pm2.4}$\\
% $\checkmark$ & $\checkmark$ & $\checkmark$ & $\times$ & $\checkmark$  & $\mathbf{13.8}_{\pm0.2}$ & $\mathbf{42.7}_{\pm1.2}$ & $\mathbf{59.0}_{\pm1.6}$ & $\mathbf{38.5}_{\pm1.0}$\\ \midrule
% $\times$ & $\times$ & $\checkmark$ & $\times$ & $\times$   & $8.7_{\pm1.6}$ & $32.4_{\pm3.1}$ & $45.3_{\pm1.0}$ & $28.8_{\pm1.2}$\\
% $\checkmark$ & $\checkmark$ & $\checkmark$ & $\times$ & $\times$   & $10.5_{\pm0.9}$ & $39.0_{\pm3.7}$ & $53.7_{\pm2.2}$ & $34.4_{\pm2.0}$\\\bottomrule
% \end{tabular}
% \end{table}

\section{Conclusion}\label{sec:conclusion}

\textbf{Limitation.} This work presents several limitations that must be acknowledged. Even though \emph{DeepTFUS} is the first deep learning model taking into account transducer geometry variability (centimeter order variation in aperture and radius of curvature), this restricted variability limits the generalization of the results across a wider range of transducer configurations (e.g., helmet-like), which would require specifically adapted datasets. Additionally, the simulations are performed at a single frequency, which does not account for the range of frequencies employed in different tFUS applications. The skull dataset, while carefully curated, remains limited in size and diversity. Importantly, it does not encompass the full morphological variability of skull anatomy across different populations, which could affect the robustness and applicability of the proposed model to broader demographic groups. Moreover, the use of pseudo-CT data generated from MR images instead of actual CT scans introduces additional uncertainty. While the MR-to-pCT model offers practical advantages and avoids radiation exposure, it may lack the fidelity of true CT data in capturing bone density variations critical for accurate acoustic modeling. More precise skull modeling, including more realistic skull acoustic parameters, could also be considered in the future to reduce uncertainties in the pressure field values. %(including a better modeling of alveoli in the trabecular bone).

\textbf{Impact and safeguards.} It is important to emphasize that the model and dataset are intended strictly for research purposes. The \emph{DeepTFUS} model and accompanying dataset are not validated for clinical decision-making and must not be used as a substitute for certified medical devices or simulation platforms. As with any simulation-based method in biomedical contexts, there remains a risk that third parties could employ these tools beyond their intended scope. To mitigate this, clear disclaimers and licensing restrictions have been implemented to underscore the non-clinical nature of the data and promote responsible use. Broader impact is expected in terms of advancing reproducibility and collaboration in scientific computing, particularly in the development of data-driven methods for neurotechnology and therapeutic ultrasound.

\textbf{Conclusion.} Transcranial focused ultrasound (tFUS) holds significant promise for non-invasive neuromodulation and therapeutic applications, yet its clinical translation is impeded by the complexities of acoustic propagation through the skull and the computational demands of subject-specific simulation. In this work, we address these challenges through two key contributions. First, we present \emph{TFUScapes}, the first large-scale, publicly available dataset of full-wave 3D transcranial ultrasound simulations, capturing anatomically realistic variability across subjects and transducer configurations. This dataset provides a critical foundation for developing and benchmarking data-driven methods in tFUS. Second, we introduce \emph{DeepTFUS}, a deep learning framework that predicts 3D acoustic pressure fields from pseudo-CT volumes and arbitrary transducer geometries. By integrating anatomical and geometric information through transducer-aware feature fusion and a task-specific loss formulation, \emph{DeepTFUS} achieves accurate, high-resolution pressure field predictions at a fraction of the computational cost of traditional solvers. Together, these contributions demonstrate the feasibility and utility of combining large-scale simulation datasets with tailored neural architectures to enable rapid and accurate acoustic field estimation. Leveraging over $750$ GPU hours for simulation and generating more than $200$ GB of simulation data, our goal is to establish a shared framework that facilitates collaborative research on deep learning methods for acoustic wave propagation, ultimately advancing innovation in neurotechnology.

%While certain limitations remain — such as the restricted diversity in skull morphologies, the fixed simulation frequency, and the use of pseudo-CT rather than actual CT data — 

\section*{Acknowledgements}
This work was partially supported by the Interdisciplinary Thematic Institute HealthTech, as part of the ITI 2021-2028 program of the University of Strasbourg, CNRS, and Inserm, funded by IdEx Unistra (ANR-10-IDEX-0002) and SFRI (STRAT’US project, ANR-20-SFRI-0012) under the framework of the French Investments for the Future Program. This work was also partially supported by French state funds managed by the ANR under Grants ANR-20-CHIA-0029-01 and ANR-10-IAHU-02. This work was granted access to the HPC resources of IDRIS under the allocations AD011011631R4 made by GENCI. The authors would like to acknowledge the High-Performance Computing Center of the University of Strasbourg for supporting this work by providing scientific support and access to computing resources. Part of the computing resources were funded by the Equipex Equip@Meso project (Programme Investissements d'Avenir) and the CPER Alsacalcul/Big Data.
{
\small
\bibliographystyle{plain}
\bibliography{main_arxiv}
}
\end{document}